\documentclass[sigconf]{acmart}
\AtBeginDocument{%
  }

\copyrightyear{2026}
\acmYear{2026}
\setcopyright{cc}
\setcctype{by}
\acmConference[DAC '26]{63rd ACM/IEEE Design Automation Conference}{July 26--29, 2026}{Long Beach, CA, USA}
\acmBooktitle{63rd ACM/IEEE Design Automation Conference (DAC '26), July 26--29, 2026, Long Beach, CA, USA}
\acmDOI{10.1145/3770743.3804296}
\acmISBN{979-8-4007-2254-7/2026/07}

\usepackage{url}
\usepackage{amsmath,amsfonts}
\usepackage{algorithm}
\usepackage{algpseudocode}
\usepackage{graphicx}
\usepackage{textcomp}
\usepackage{xcolor}
\usepackage{xspace}
\usepackage{comment}
\usepackage{kotex}
\usepackage{enumitem}
\usepackage{booktabs}
\usepackage{multirow}
\usepackage{makecell}
\usepackage[normalem]{ulem}
\usepackage{soul}
\usepackage{placeins}
\usepackage{float}

\newcommand{\SysName}{CoX-MoE\xspace}
\title{\SysName: Coalesced Expert Execution for High-Throughput MoE Inference with AMX-Enabled CPU–GPU Co-Execution}

\author{Muyoung Son}
\authornote{Both authors contributed equally to this research.}
\affiliation{%
  \institution{KAIST}
  \city{Daejeon}
  \country{Republic of Korea}
}
\email{kkt1690@kaist.ac.kr}

\author{Yi Chen}
\authornotemark[1]
\affiliation{%
  \institution{KAIST}
  \city{Daejeon}
  \country{Republic of Korea}
}
\email{chenyi@kaist.ac.kr}

\author{Seungjae Yoo}
\affiliation{%
  \institution{KAIST}
  \city{Daejeon}
  \country{Republic of Korea}
}
\email{goldenyoo@kaist.ac.kr}

\author{Soongyu Choi}
\affiliation{%
  \institution{KAIST}
  \city{Daejeon}
  \country{Republic of Korea}
}
\email{soongyu1291@kaist.ac.kr}

\author{Joo-Young Kim}
\authornote{Corresponding author.}
\affiliation{%
  \institution{KAIST}
  \city{Daejeon}
  \country{Republic of Korea}
}
\email{jooyoung1203@kaist.ac.kr}

\ccsdesc[500]{Computer systems organization~Heterogeneous (hybrid) systems}
\ccsdesc[500]{Computing methodologies~Natural language processing}
\begin{document}
\begin{abstract}
The Mixture-of-Experts (MoE) architecture improves computational efficiency via sparse expert activation, but throughput-oriented inference faces substantial GPU memory pressure due to a significant parameter size and intermediate data. Prior works attempt to mitigate this using expert offloading with micro-batching or by offloading computation to the CPU. However, the fragmented workload resulting from micro-batching degrades operational intensity, causing expert execution to become memory-bound. Meanwhile, CPU offloading is constrained by slow PCIe transfers and its limited applicability to attention computation in the decode stage. Consequently, these inefficiencies prevent effective system utilization, severely restricting the end-to-end throughput of MoE inference.

To address these challenges, this paper proposes \SysName, an Advanced Matrix Extensions (AMX)–enabled CPU–GPU collaborative system that comprehensively optimizes MoE inference by combining coalesced expert execution with strategic workload orchestration for higher throughput. \SysName introduces (i) a coalescing-aware orchestration policy to jointly optimize resource allocation by adopting ordinary batch, instead of micro-batch, for expert computation and selective attention offloading, and (ii) a static expert-aware stratification scheme that pre-assigns frequently activated experts to the GPU, mitigating PCIe transfer overhead and balancing workload for the CPU and GPU during inference. Compared to state-of-the-art frameworks, \SysName delivers significant gains, achieving up to 7.1$\times$ and 2.4$\times$ higher throughput than FlexGen and MoE-Lightning, respectively.
\end{abstract}

\keywords{MoE Inference, Offloading, Workload Orchestration, Heterogeneous Computing, AMX}
\maketitle


\begin{figure}[t]
    \Description{A block diagram illustrating the Mixture-of-Experts (MoE) architecture, showing tokens being routed by a router to sparse experts, along with the overall inference data flow.}
    \centering
    \includegraphics[width=0.90\linewidth]{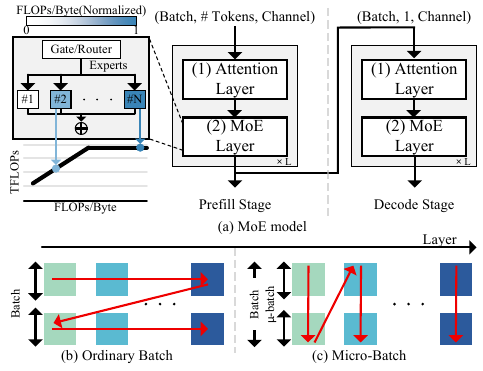}
    \vspace{-14pt}
    \caption{(a) Architecture of Mixture-of-Experts along with the inference flow. (b), (c) Inference flow for each strategy.}
    \vspace{-14pt}
    \label{fig:bg_1}
\end{figure}

\section{Introduction}
\label{section1}
\vspace{-4pt}


Mixture-of-Experts (MoE) has emerged as a promising architecture for scaling model capacity without proportionally increasing computation cost ~\citep{zhou2022mixture, liu2024survey}. Unlike dense large language models (LLMs)~\citep{chang2024survey,naveed2025comprehensive}, which activate all parameters, MoE activates only a subset of experts for each token, reducing active parameters during inference while maintaining high performance. As a result, MoEs are considered more computationally efficient than dense LLMs.

Despite its advantage, the MoE architecture poses a severe GPU memory (VRAM) challenge. MoE models typically have larger parameter sizes than dense LLMs, despite having similar active parameters. Moreover, throughput-oriented workloads, such as offline benchmarking~\citep{liang2022holistic}, large-scale data processing~\citep{narayan2018don}, or synthetic data generation~\citep{lu2023machine}, further inflate VRAM usage due to large intermediate activations. For example, Mixtral-8x22B~\citep{jiang2024mixtral} requires about 282 GB for BF16 weights, and a 64-batch, 4096-token workload adds roughly 72 GB of intermediate data, bringing the total to around 350 GB. Such footprints require multiple high-end GPUs (e.g., roughly five 80 GB H100s~\citep{choquette2023nvidia}, costing up to $\$200,000$), which is impractical for most deployments and motivates MoE inference systems that focus on single-GPU, throughput-intensive workloads.

To operate high-throughput inference within a limited VRAM capacity, prior works~\citep{sheng2023flexgen, cao2025moe} combine memory offloading with micro-batching. The memory offloading strategy involves offloading model weights or activation data into host memory and swapping them between VRAM and host memory via PCIe, enabling inference for models that exceed VRAM capacity. For MoE models, memory offloading takes the form of expert offloading~\citep{tang2024hobbit,jiang2024mixtral,xue2024moe}, where non-expert and a subset of expert weights remain resident in VRAM, while other expert weights are fetched from host memory via on-demand~\citep{tang2024hobbit} or prefetch~\citep{hwang2024pre, zhong2024adapmoe, tang2024hobbit} mechanisms. To maximize the efficiency of memory offloading, large input batches are divided into multiple micro-batches, which are executed sequentially on the GPU, since activations for the entire batch often exceed VRAM capacity. As shown in Fig.~\ref{fig:bg_1}(c), this design promotes weight reuse across micro-batches and alleviates memory pressure. In alternative approaches, other systems~\citep{kamahori2024fiddler, zhong2025hybrimoe, chen2025ktransformers} offload expert operations to the CPU to reduce PCIe transfer, which is one of the primary bottlenecks, and examine the activated experts at each step to decide whether each should be executed on the GPU or the CPU.

However, these systems face two limitations for MoE inference. First, micro-batching fragments each expert’s workload, lowering its operational intensity and causing expert computations to become memory-bound, where latency is dominated by VRAM access. This bottleneck is then exacerbated by repeatedly loading the same expert weights from VRAM in every micro-batch, resulting in increased per-layer latency and reduced overall throughput. 

Second, existing CPU-assist solutions primarily rely on Intel Advanced Vector Extensions (AVX)~\citep{intel_dl_boost_guide} and target offloading attention-related GEMV operations in the decode stage, while leaving the GEMM-intensive prefill stage largely unexploited. The reason is that the limited per-core matrix multiplication throughput of AVX is insufficient to enable meaningful compute offloading. Furthermore, attempting to offload expert computation is hampered by fundamental hardware limitations: the slow PCIe connection, and the significant performance gap between the CPU and GPU may lead to workload imbalance, thereby limiting the overall performance.

To overcome these challenges, we propose \SysName, an Advanced Matrix Extensions (AMX)-enabled CPU–GPU collaborative system that creates an optimization framework centering on (i) coalescing-aware orchestration policy and (ii) expert-aware stratification to maximize MoE inference throughput. In summary, we make the following contributions:

\begin{itemize}[leftmargin=*]
    \item We analyze the workload of MoE inference, identifying key performance bottlenecks and optimization opportunities by leveraging modern CPU instructions with Intel AMX.
    \item We propose and implement a coalescing-aware orchestration policy implemented within an efficient AMX-enabled CPU-GPU framework, which jointly optimizes compute/expert allocation, while enforcing ordinary batch for expert computation instead of micro-batch, and utilizing an attention offloading strategy that frees up VRAM by assigning partial or complete attention operations to the CPU or GPU.
    \item We design an expert-aware stratification scheme to identify frequently activated experts, based on batch clustering, sampling and probing, enabling ahead-of-time expert placement that maximizes resource utilization and balances the workload.
    \item Overall, \SysName achieves a 1.7–2.4$\times$ throughput improvement over a state-of-the-art (SOTA) offloading scheme.
\end{itemize}

\section{Background}
\label{section2}
\vspace{-4pt}

\subsection{MoE Models and Inference Workloads}
\label{section2_subsec1}
\vspace{-2pt}
As shown in Fig.~\ref{fig:bg_1}(a), MoE models share the same two-phase inference with dense LLMs~\citep{hong2022dfx}. The prefill stage processes the entire batch of input sequences at once. In contrast, the decode stage generates tokens autoregressively, one token at a time. While dense LLMs activate all parameters for every input token, MoE models employ a routing function to dynamically route each token to a sparse subset of experts, for example, selecting the top-k from a much larger pool of $N$  total experts~\citep{chen2022towards}. This sparse and uneven token-to-expert routing results in varying workloads across individual experts, leading to substantial variation in operational intensity.

\vspace{-6pt}
\subsection{Advanced Matrix Extensions}
\label{section2_subsec2}
\vspace{-2pt}
To accelerate CPU-side inference for ML workloads, Intel introduced AMX, an on-chip matrix multiplication accelerator with dedicated ISA support, starting with the 4th generation Xeon (2022)~\citep{briefaccelerate, kim2024exploiting}. The AMX architecture comprises two key components: (1) a 2D array of registers (Tile) and (2) a Tile Matrix-Multiply (TMUL) unit, both designed to operate efficiently on INT8 and BF16 data formats. To deliver high inference speed, the CPU issues dedicated AMX instructions which execute on multi-cycle AMX units.

Compared to AVX-512~\citep{intel_dl_boost_guide}, which delivers approximately 18 TFLOPs of peak BF16 performance per socket, Intel AMX on Sapphire Rapids increases per-socket matrix multiplication throughput to approximately 144 TFLOPs, about an order of magnitude higher. This substantial improvement narrows the gap to a high-end GPU, such as the RTX 6000 Ada, at approximately 364 TFLOPs. This elevated CPU-side throughput with AMX makes CPU-GPU co-execution a practically effective strategy for compute offloading.

\vspace{-6pt}
\subsection{Offloading and Micro-Batch Strategy}
\label{section2_subsec3}
\vspace{-2pt}
Traditional inference systems face significant VRAM pressure when model size exceeds memory or when processing large batches. To address this, unlike normal batch inference shown in Fig.~\ref{fig:bg_1}(b), prior works~\citep{sheng2023flexgen, cao2025moe} offload model weights and activations to host memory or SSDs and process each batch as a sequence of micro-batches on the GPU, as illustrated in Fig.~\ref{fig:bg_1}(c). By executing these micro-batches sequentially, the system enables weight reuse: the expert weights are loaded into VRAM once and reused across all micro-batches, significantly reducing repeated PCIe transfers. The micro-batch size is optimized to balance between PCIe transfer time and on-GPU efficiency. For MoE models, a specialized mechanism known as expert offloading is used~\citep{zhong2024adapmoe,xue2024moe,tang2024hobbit}. Non-expert and a subset of expert weights remain resident on the GPU, while the other experts are stored in host memory or on an SSD. When activated, these offloaded experts are either fetched to the GPU (via prefetch or on-demand loading), or computed directly on the CPU~\citep{kamahori2024fiddler, chen2025ktransformers, zhong2025hybrimoe}.

Furthermore, in the decode stage, because the KV cache grows dynamically and makes attention heavily memory-bound, repeatedly transferring it over the slow PCIe bus is inefficient. Therefore, prior works~\citep{sheng2023flexgen,cao2025moe} also fully offload decode-stage attention to the CPU to operate directly on KV data in host memory.

However, these approaches remain limited by the low operational intensity of expert execution under micro-batching and the insufficient CPU throughput in compute offloading. Furthermore, existing works suffer from severe workload imbalance when offloading experts and have a narrow focus on decode-stage attention.

\section{Motivation}
\label{section3}
\vspace{-4pt}

\begin{figure}[t]
    \Description{A graph analyzing the impact of micro-batch size on different components of MoE inference latency, showing that expert computation is the dominant component and its latency increases linearly as micro-batch size decreases.}
    \centering
    \includegraphics[width=0.9\linewidth]{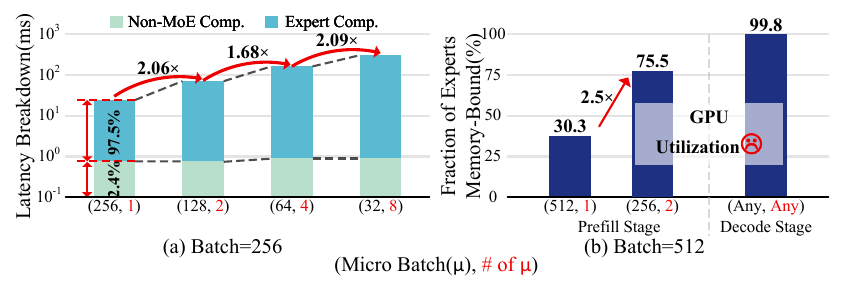}
    \vspace{-14pt}
    \caption{Analysis of Micro-Batching Strategy for Inference.}
    \vspace{-14pt}
    \label{fig:mov_1}
\end{figure}



In this section, we use NVIDIA Nsight Systems/Compute~\citep{nvidia_nsight} to profile the Qwen3-30B-A3B~\citep{yang2025qwen3} on an NVIDIA RTX 6000 Ada Generation GPU with an Intel Xeon Platinum 8452Y CPU, identifying performance bottlenecks and characterizing MoE batch inference behavior. We evaluate batch sizes of 128, 256, and 512 with a prompt length of 512, following prior work~\citep{patel2024splitwise, zhong2025hybrimoe}.

\subsection{Low Expert Arithmetic Intensity}
\label{section3_subsec1}
\vspace{-2pt}

As shown in Fig.~\ref{fig:mov_1}(a), E2E latency increases with the number of micro-batches, while non-expert latency remains nearly constant. In contrast, expert computation, which accounts for over 97.5\% of total latency, is highly sensitive to micro-batching: halving the micro-batch size nearly doubles expert latency. This is because sparse routing gives each expert fewer tokens than non-expert operations, and smaller micro-batches further reduce per-expert inputs, lowering operational intensity and pushing expert execution into the memory-bound regime. Accordingly, Fig.~\ref{fig:mov_1}(b) shows that more experts become memory-bound as micro-batch size decreases.

Once an expert becomes memory-bound, its execution time is dominated by repeated fetches of expert weights from VRAM per micro-batch, which in turn increases the total VRAM access cost and directly leads to a near-linear increase in expert latency. In contrast, non-expert computation remains stable, as it has not yet entered the memory-bound regime under the evaluated micro-batch sizes.
Consequently, prior work that relies on micro-batching dramatically exacerbates the primary latency bottleneck by inflating expert computation time, severely limiting overall throughput.


\textbf{Opportunity 1: Coalesced Expert Execution with AMX-Assisted Co-Execution}. The severe penalty of micro-batching reveals a key opportunity: execute experts over the entire batch instead of splitting them into micro-batches. This coalesced execution increases operational intensity, but also concentrates computation on the GPU. Meanwhile, as discussed in Section~\ref{section2_subsec2}, AMX provides high enough performance for the CPU to offload a meaningful portion of expert computation. These observations motivate AMX-enabled CPU-GPU co-execution for higher throughput.


\vspace{-6pt}
\subsection{Intermediate Data-driven VRAM Pressure}
\label{section3_subsec2}
\begin{figure}[t]
    \Description{A multi-part figure showing: (a) VRAM partitioning comparison between GPU and CPU attention offloading with resulting latency reduction, (b) A Roofline model comparing GPU, AMX, and PCIe bandwidth limits for MoE operations, and (c) The heavily skewed workload distribution across different expert indices.}
    \centering
    \includegraphics[width=\linewidth]{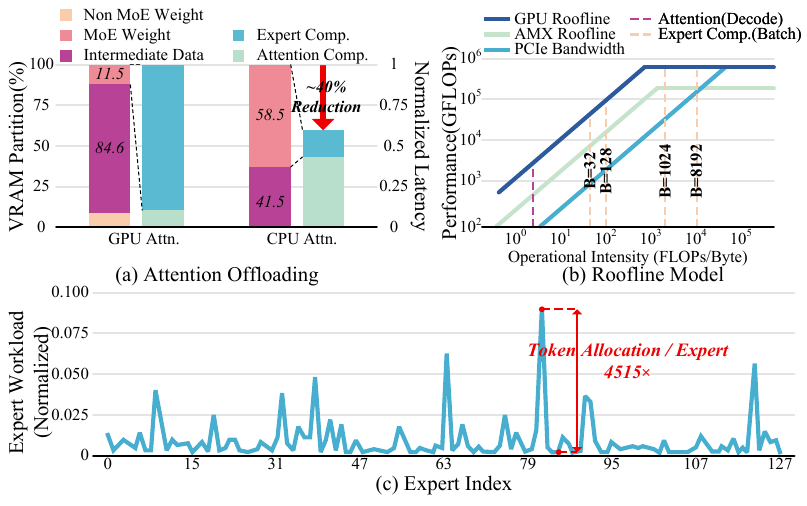}
    \vspace{-14pt}
    \caption{(a) VRAM partitioning and Latency analysis via Attention Offloading. (b) GPU and AMX Roofline Model with PCIe. (c) Normalized expert workload distribution at Layer 1 on the BIG-Bench Hard (bbh) benchmark.}
    \vspace{-14pt}
    \label{fig:mov_2}
\end{figure}

\vspace{-2pt}
Fig.~\ref{fig:mov_2}(a) illustrates the breakdown of VRAM consumption during MoE inference, highlighting the substantial memory pressure imposed by attention operations in the prefill stage. In the baseline ``GPU Attn'', which executes attention operations on the GPU, the vast majority of VRAM (84.6\%) is consumed by intermediate data generated during computation. It severely constrains the VRAM available for expert weights, limiting them to only 11.5\% of the total partition. Consequently, most experts must be offloaded from the GPU, forcing either frequent data transfers over the PCIe bus or imposing a heavy computational load on the CPU. This memory constraint severely exacerbates the expert computation bottleneck.

\textbf{Opportunity 2: Optimal VRAM Allocation via Attention Offloading}. We analyze a ``CPU Attn'' that offloads the attention computation from the GPU to the CPU using AMX, allowing more expert weights to be stored in VRAM, expanding to 58.5\%. This strategy introduces a clear trade-off: the attention computation latency increases because data is moved into the CPU. However, keeping more experts resident in VRAM dramatically reduces expert computation latency. 
As shown in Fig.~\ref{fig:mov_2}(a), this reduction exceeds the additional attention delay, with AMX providing sufficient CPU-side GEMM capability, resulting in a ~40\% reduction in total inference latency.
This finding implies that for large-batch MoE model inference, it is more effective to strategically offload attention computation, which generates the voluminous intermediate data, rather than offloading expert weights.


\vspace{-6pt}
\subsection{PCIe Bottleneck and Workload Imbalance}
\label{section3_subsec3}
\vspace{-2pt}
As shown in Fig.~\ref{fig:mov_2}(b), offloading experts and their computations introduces two critical performance bottlenecks. The first is the PCIe bandwidth bottleneck, where the PCIe transfer rate ($\approx 32$ GB/s) for fetching experts to the GPU is 10$\times$ slower than the CPU's DDR5 memory bandwidth ($\approx 300$ GB/s). This significant difference ensures that dynamically transferring experts over the PCIe bus results in significant I/O overhead.
The second is the asymmetry in computational power. As discussed in Section~\ref{section3_subsec1}, Intel AMX significantly boosts CPU throughput, making co-execution feasible. However, a performance gap with the GPU still remains. This means that simply offloading experts without considering their workload intensity can lead to a workload imbalance, shifting the bottleneck from the GPU to the CPU. Therefore, an effective co-execution policy must carefully balance this asymmetry by assigning the appropriate workload to each device.

\textbf{Opportunity 3: Static Allocation of Skewed Experts into VRAM}. Our empirical analysis reveals that MoE architectures are characterized by sparse and uneven activated experts across all layers, as exemplified by Fig.~\ref{fig:mov_2}(c): usage frequency varies drastically among experts, indicating that not all experts contribute equally to the computational workload. This skewed expert activation pattern suggests an opportunity to statically prioritize frequently used experts in VRAM, which is crucial for minimizing PCIe overhead and balancing the overall system throughput.


\begin{figure}[t]
    \Description{An illustrative diagram detailing the proposed MoE offloading policy, showing how different components (Experts, Attention, etc.) are strategically partitioned between the GPU and CPU for computation and storage.}
    \centering
    \includegraphics[width=0.9\linewidth]{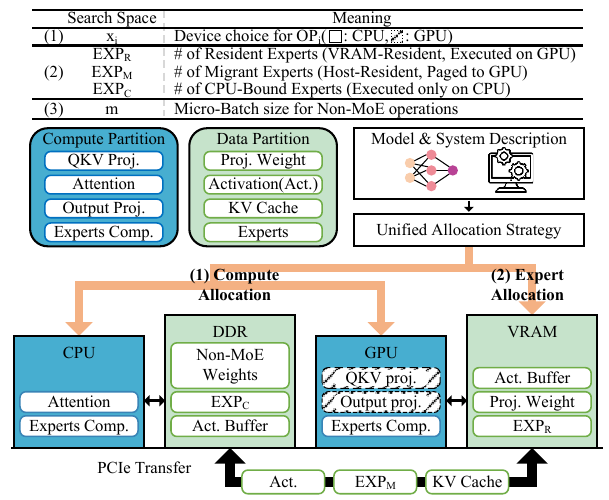}
    \vspace{-14pt}
    \caption{Unified Allocation Strategy.}
    \vspace{-14pt}
    \label{fig:feature_1}
\end{figure}

\section{\SysName Design}
\label{section4}
\vspace{-4pt}

\subsection{Coalescing-Aware Orchestration Policy: Micro-Batch is not all you need.}
\vspace{-2pt}
\begin{table}[t]
  \centering
  \caption{Per–micro-batch (\(m\)) data size and FLOPs of decoder operations (BF16).}
  \vspace{-10pt}
  \label{tab:bytes_flops}
  \small
  \renewcommand{\arraystretch}{1.1}
  \scalebox{0.97}{ 
  \begin{tabular}{@{}c l c c c@{}}
    \toprule
    \textbf{Idx(i)} & \textbf{Operation(\(OP_i\))} & \(D_X\) (Byte) & \(D_Y\) (Byte) & \(C\) (FLOP) \\
    \midrule
    0 & QKV Projection & \(2mL d_h\) & \(6d_h^2\) & \(6mLd_h^2\) \\
    1 & Attention(Prefill) & \(2mL d_h\) & \(4mL d_h\) & \(4mL^2 d_h\) \\
      & Attention(Decode) & & \(4mL_{kv} d_h\) & \(4mL_{kv}d_h\) \\
    2 & Out. Projection & \(2mL d_h\) & \(2d_h^2\) & \(2mL d_h^2\) \\
    3 & Per-Expert Computation  & \(\frac{2BLd_h}{E}\) & \(6d_hd_e\) & \(\frac{6BLd_hd_e}{E}\) \\
    \bottomrule
  \end{tabular}
  }
  \\[1pt]  
  \raggedright\footnotesize
  \(B\): Batch size, \(E\): Number of activated experts, \(d_h\): model (hidden) dim, \(d_e\): expert dim, \(L\): sequence length (equals 1 in the decode stage), \(L_{kv}\): decode stage KV length. 
\vspace{-10pt}
\end{table}

\SysName establishes an optimal offloading policy based on computational characteristics to minimize e2e latency during the MoE model inference in a CPU-GPU system. First, \textit{Unified Allocation Strategy}, as shown in Fig.~\ref{fig:feature_1}, jointly optimizes two coupled decisions: (1) compute allocation for operations and (2) expert allocation to VRAM. Second, with \textit{Strategy-aware Micro-batch Determination}, \SysName selects the micro-batch size \(m\) for non-MoE operations (implying the number of micro-batches $M = \lceil B/m \rceil$), while fixing expert computation to use a coalesced batch of size \(B\). These twp-step decisions are established to minimize per-layer latency ($T_{l}$) to achieve the optimum ($T_{opt}$), thereby reducing the total latency ($T_{tot}$) for the entire model.

\begin{equation}\label{eq:obj}
T_{tot} = \sum_{l=1}^{N} T_l \ge N \times \min(T_{l}) = N \times T_{opt}
\end{equation}
Here, $N$ is the total number of decoder layers, and $T_l$ represents the latency of the $l$-th decoder layer.
As shown in Table~\ref{tab:bytes_flops}, \SysName partitions a decoder layer into four operation units, $OP_0$ (QKV projection), $OP_1$ (attention), $OP_2$ (output projection), and $OP_3$ (FFN of experts), which represent Q/K/V formation via linear maps, attention with KV-cache access, projection to the model hidden size, and gated FFN execution per expert, respectively. In addition, Table~\ref{tab:bytes_flops} summarizes the data size of the first and second matrix operands (\(D_{X_i}\), \(D_{Y_i}\)) and FLOP(\(C_{i}\)) in each operation, with quantities expressed per $m$ or per $B$ as noted.

\begin{figure}[t]
    \Description{A timing diagram illustrating the concurrent execution and pipeline stages of MoE operations within a single layer across the GPU and CPU, highlighting the overlapping of computation and data transfer to maximize throughput.}
    \centering
    \includegraphics[width=0.9\linewidth]{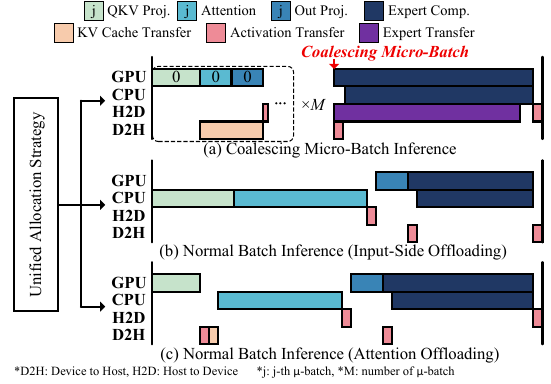}
    \vspace{-14pt}
    \caption{Example of timing diagram for a single layer of Qwen3-30B-A3B. Note that block sizes do not represent actual latency.}
    \vspace{-14pt}
    \label{fig:feature_2}
\end{figure}
\vspace{-4pt}
\subsubsection{\textbf{Unified Allocation Strategy}}
\vspace{-2pt}
As shown in Fig.~\ref{fig:feature_1}, \SysName jointly optimizes two coupled decisions. First, as mentioned in Section~\ref{section3_subsec2}, it assigns each of the three non-MoE operations exclusively to either the CPU or the GPU. Second, as mentioned in Section~\ref{section3_subsec1}, it prioritizes coalesced expert execution—processing the entire batch $B$ as a single unit—while enabling collaborative execution between the CPU and GPU.

The optimal single-layer latency ($T_{opt}$) aims to minimize the sum of the latencies of operational units.
\begin{equation}\label{eq:t_opt}
T_{opt} = \min \left( \sum_{i=0}^3 (T(OP_i)) \right)
\end{equation}
The latency of each operation ($T(OP_i)$, where $i \in \{0, 1, 2, 3\}$) is modeled as the sum of its loading time the two operand matrices in a given operation ($T_{\text{load}}$), the computation time ($T_{\text{comp}}$) and KV cache store time ($T_{\text{store}}$).
\begin{equation}\label{eq:t_op_i}
T(OP_i)=T_{\text{load}}(OP_i)+T_{\text{comp}}(OP_i)+T_{\text{store}}(OP_i)
\end{equation}
$T_{\text{load}}$ accounts for PCIe transfers of activations when the execution device changes between consecutive stages.

\begin{equation}\label{eq:t_op_load}
T_{\text{load}}(OP_i)=
\begin{cases}
M D_{X_i} / BW_{\mathrm{PCIe}}, & \text{if } x_i \ne x_{i-1},\\
0, & \text{otherwise.}
\end{cases}
\end{equation}

$T_{\text{comp}}$ follows a roofline model based on the hardware configuration of each device. The latency is determined by the bottleneck resource, which is either the memory bandwidth (\(BW\)) or the computational performance (\(TF\)).

\vspace{-8pt}
\begin{equation}\label{eq:t_op_comp}
T_{\text{comp}}(OP_i)
= M \max \!\left(
(D_{X_i}+D_{Y_i})/BW_{\mathrm{DEV}},
\ \mathrm{C}_i/\mathrm{TF}_{\mathrm{DEV}}
\right)
\end{equation}

Finally, $T_{\text{store}}$ is 
\begin{equation}\label{eq:t_op_store}
T_{\text{store}}(OP_i)=
\begin{cases}
M D_{\mathrm{KV}} / BW_{\mathrm{PCIe}},
& \text{if } i=1,\ x_{0}=1,\ x_{1}=0,\\
0, & \text{otherwise.}
\end{cases}
\end{equation}
where \(D_{KV}\) is the per micro-batch KV cache size generated from $OP_0$. The transfer via PCIe is incurred precisely when attention runs on the CPU ($x_{1}=0$) 
while QKV runs on the GPU ($x_{0}=1$).

With an allocation exclusively to $ OP_0$-$OP_2$, the optimization focus shifts to $OP_3$. Under VRAM capacity and PCIe bandwidth constraints, the number of experts to process on each device should be determined. We classify experts into three groups: $EXP_{\text{R}}$, $EXP_{\text{M}}$, and $EXP_{\text{C}}$, as depicted at Fig.~\ref{fig:feature_1}. 
As in Eq.~\eqref{eq:t_op_i}, we now specialize the per-operation latency model to the expert stage $OP_3$ by assuming no micro-batch ($m_{exp}=B$) and parallel CPU-GPU execution.


\begin{equation}\label{eq:t_op_load_3}
T_{\text{load}}(OP_3)
= (D_{X_3} + EXP_M D_{Y_3}) / BW_{\mathrm{PCIe}}
\end{equation}
\vspace{-20pt}

\begin{equation}\label{eq:t_op_comp_3}
T_{\text{comp}}(OP_3)
= \max(\mathrm{Latency}_{\mathrm{GPU}}, \mathrm{Latency}_{\mathrm{CPU}})
\end{equation}
\vspace{-20pt}

\begin{equation}\label{eq:latency_dev}
\mathrm{Latency}_{\mathrm{DEV}}
= \max \bigl(
(D_{X_3} + EXP_{\mathrm{DEV}} D_{Y_3}) / BW_{\mathrm{DEV}},
\ C_3 / TF_{\mathrm{DEV}}
\bigr)
\end{equation}

Where $\mathrm{DEV}\in\{\mathrm{CPU},\mathrm{GPU}\}$, with
$EXP_{\mathrm{DEV}}\in\{EXP_{\mathrm{R}}{+}\allowbreak EXP_{\mathrm{M}},\,\allowbreak EXP_{\mathrm{C}}\}$.
This ties $(EXP_{\mathrm{R}},EXP_{\mathrm{M}},EXP_{\mathrm{C}})$ directly
to $T(\mathrm{OP}_3)$ and thus to $T_{\mathrm{opt}}$.
As shown in Table~\ref{tab:bytes_flops}, $OP_3$ uses an ordinary batch \(B\) for $D_x$, effectively setting the micro-batch size for this stage to \(B\).

\vspace{-4pt}
\subsubsection{\textbf{Strategy-Aware Micro-Batch Determination.}}
\vspace{-2pt}
Given the above strategy, \SysName determines the optimal \(m\) for $OP_0$–$OP_2$ that satisfies Eq.~\ref{eq:obj}. under the VRAM budget. The budget explicitly accounts for $EXP_{\text{R}}$ weights, computes allocation-induced non-MoE weights, intermediate data buffers (for the attention/expert computation), and temporary working space for kernels. \SysName then searches the feasible space of \((x_0,x_1,x_2, EXP_{\mathrm{R}}, EXP_{\mathrm{M}}, EXP_{\mathrm{C}},m)\) that satisfies the VRAM/PCIe constraints and selects the configuration that yields the smallest \(T_{\mathrm{tot}}\), thereby maximizing throughput.

\vspace{-4pt}
\subsection{Performance Optimization}
\vspace{-2pt}
\begin{figure}[t]
    \Description{}
    \centering
    \includegraphics[width=0.9\linewidth]{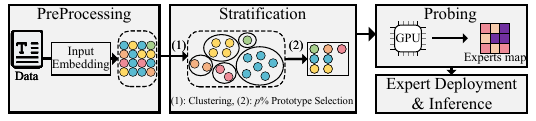}
    \vspace{-14pt}
    \caption{Expert-Aware Stratification Workflow.}
    \vspace{-14pt}
    \label{fig:feature_3}
\end{figure}
Once the two-step decisions are determined, the execution strategy is defined. Fig.~\ref{fig:feature_2} illustrates a timing diagram across prefill and decode with two primary strategies. First, in Fig.~\ref{fig:feature_2}(a), the $OP_0$-$OP_2$ are performed in micro-batch units, but the expert computation is performed after coalescing these micro-batches. The key optimization is overlapping data movement between the CPU and GPU with computation to remove idle gaps. The second, as exemplified by Fig.~\ref{fig:feature_2}(b) and (c), the system adopts a normal batch inference for the entire layer. Since PCIe transfer latency tends to dominate computation, especially during the decode stage, this strategy prioritizes minimizing the transfer of expert weights between CPU and GPU.

\vspace{-4pt}
\subsection{Expert-Aware Stratification}
\vspace{-2pt}
As mentioned in Section~\ref {section3_subsec3}, \SysName introduces Expert-Aware Stratification (EAS), a lightweight data-driven pre-analysis framework that selects which experts should be statically preloaded into VRAM before inference, thereby minimizing dynamic PCIe transfer overhead and workload imbalance between GPU and CPU. EAS is particularly advantageous in batch inference scenarios, such as throughput-oriented benchmarking, where the entire inference workload is available a priori. In these scenarios, profiling the entire dataset is prohibitively expensive. Thus, the key challenge is to identify a representative subset of data that faithfully reflects the global expert activation pattern.

As shown in Fig.~\ref{fig:feature_3}, EAS operates in three steps. First, in the pre-processing stage, it generates input embeddings for the entire dataset, capturing the latent semantic distribution of inputs. Second, in the stratification stage, the embeddings are clustered to form strata that group semantically similar samples. The representative prototypes are then proportionally selected per cluster, ensuring stratified coverage of the dataset. Finally, inference is performed only on these selected prototypes during a prefill-only probing phase, from which the global expert activation map is approximated. The approximated expert activation map guides static expert deployment, allowing frequently activated experts to be initialized on the GPU while keeping low-usage experts offloaded to the CPU.

\section{System Implementation}
\label{section5}
\vspace{-4pt}

To enable the efficient execution of MoE models on AMX-enabled CPU-GPU systems, we extend the Intel Extension for PyTorch (IPEX), which lacks native NVIDIA support, to facilitate interoperability with NVIDIA GPUs. Moreover, we implement fine-grained, globally shared CUDA streams that allow parallel CPU-GPU execution and VRAM buffer sharing. This design maximizes throughput by overlapping PCIe data transfers with computation, thus ensuring optimal system resource utilization.



\begin{figure*}[htbp]
    \Description{An evaluation result illustrating the throughput versus the hardware in Table.~\ref{tab:eval-system} and generation configuration for the models Mixtral-8x7B, DeepSeek-V2-Lite and Qwen3-30B-A3B.}
  \centering
  \includegraphics[width=0.85\textwidth]{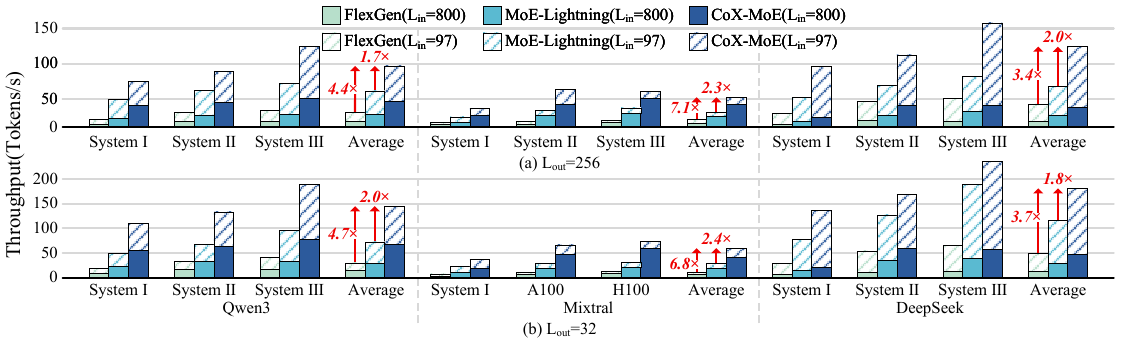}
  \vspace{-14pt}
  \caption{Inference throughput (Tokens/s) comparison between \SysName, MoE-Lightning and FlexGen acorss the system configurations in Table.~\ref{tab:eval-system}, with $B=1024$}
  \vspace{-14pt}
  \label{fig:eval_1}
\end{figure*}

\section{Experimental Results}
\label{section6}
\vspace{-4pt}

\begin{table}[t]
\centering
\caption{Evaluation Systems}
\vspace{-10pt}
  \label{tab:eval-system}
  \small
  \scalebox{0.97}{
  \begin{tabular}{@{}l@{}}
    \toprule
    \textbf{Host(CPU) Platform} \\
    Intel\textsuperscript{\textregistered} Xeon\textsuperscript{\textregistered} Platinum 8452Y, 36 cores \\
    8 DDR5-4800 channels, \textbf{512\,GB} DRAM \\
    \midrule
    \textbf{System Configurations} \\
    (I) CPU + NVIDIA 6000 Ada Generation Graphics Card, \textbf{48\,GB}, PCIe~4.0 \\
    (II) CPU + NVIDIA A100 Graphics Card, \textbf{80\,GB}, PCIe~4.0 \\
    (III) CPU + NVIDIA H100 Graphics Card, \textbf{80\,GB}, PCIe~5.0 \\
    \bottomrule
  \end{tabular}
  }
\vspace{-10pt}
\end{table}

\subsection{Experimental Setup}
\label{section6_subsec1}
\vspace{-2pt}

\textbf{Platforms}.
As shown in \autoref{tab:eval-system}, we evaluate \SysName on a 36-core Intel\textsuperscript{\textregistered} Xeon\textsuperscript{\textregistered} Platinum 8452Y with AMX~\citep{briefaccelerate} support paired with A100~\citep{choquette2020nvidia}, H100~\citep{choquette2023nvidia}, and RTX 6000 Ada Generation GPUs.

\textbf{Models}.
We evaluate our system using three representative MoE models with distinct characteristics: Mixtral-8x7B-Instruct (Mixtral)~\citep{jiang2024mixtral}, DeepSeek-V2-Lite (DeepSeek)~\citep{liu2024deepseek}, and Qwen3-30B-A3B (Qwen3)~\citep{yang2025qwen3}. These models differ in the number and size of experts, as well as their architectural configurations. Mixtral employs fewer, larger experts, in contrast to the numerous smaller experts in Qwen3 and DeepSeek.

\textbf{Baseline}.
We evaluate \SysName against two prominent open-source inference frameworks: FlexGen~\citep{sheng2023flexgen} and MoE-Lightning~\citep{cao2025moe}. FlexGen introduces a basic offloading policy that searches the compute schedule for micro-batches and executes a zig-zag block schedule. MoE-Lightning is the SOTA batch inference system via CPU-GPU-I/O (CGO) pipelining with paged expert weights and a hierarchical roofline model to select micro-batch. In both frameworks, attention in the decode stage is executed on the CPU.

\begin{figure}[t]
    \Description{An evaluation result illustrating the expert hit ratio versus stratification number of expert for DeepSeek and Qwen3, with different sampling ratios (5\% and 10\%).}
  \centering
  \includegraphics[width=0.9\columnwidth]{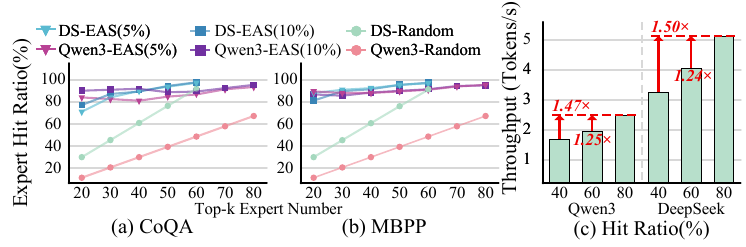}
  \vspace{-14pt}
  \caption{(a), (b) The relation ship between expert hit ratio and number of stratified experts. (c) Evaluation of throughput (Tokens/s) versus hit ratio.}
  \vspace{-12pt}
  \label{fig:eval_2}
\end{figure}

\begin{table}[t]
\centering
\caption{Breakdown of proposed techniques. Baseline is MoE-Lightning with $B=512$, $L_{\text{in}}=1320$, $L_{\text{out}}=128$, evaluated on System(I). (a) coalescing micro-batch and co-execution for experts with AMX, (b) attention offloading into CPU, (c) 80\% expert hit ratio.}
\vspace{-4pt}
\label{tab:ablation_study}
\small
\scalebox{0.97}{
\begin{tabular}{lccc}
    \hline
    Technique & \(m\) & Throughput (Tokens/s) & Improvement \\
    \hline
    Baseline & 256 & 16.8 & --        \\
    +(a) & 256 & 25.5 & 1.51$\times$ \\
    +(a)+(b) & 512 & 32.3 & 1.26$\times$ \\
    +(a)+(b)+(c) & 512 & 34.1 & 1.05$\times$ \\
    \hline
    \end{tabular}
}
\vspace{-10pt}
\end{table}


\vspace{-6pt}
\subsection{Performance Comparison}
\vspace{-2pt}
Fig.~\ref{fig:eval_1} reports the throughput for batch size 1024 under two input lengths, ($L_{in}=97$ and $800$), and two output lengths, ($L_{out}=32$ and $256$). \SysName demonstrates consistent improvements over the baselines across all $L_{in}$, $L_{out}$, models and systems. \SysName delivers 1.7--2.4$\times$ and 3.4--7.1$\times$ higher throughput compared to MoE-Lightning and FlexGen, respectively. Whether the workload is prefill-dominant (driven by large $L_{in}$) or decode-dominant (driven by large $L_{out}$), \SysName maintains a significant throughput advantage over both baselines and stages. This impact is driven by \SysName's coalescing-aware orchestration policy, which determines the strategies for compute and expert allocation to balance workloads and achieve efficient inference.
In particular, the most significant improvements are observed in the Mixtral model, where throughput increases by up to 2.4$\times$ over MoE-Lightning and 7.1$\times$ over FlexGen. Because Mixtral’s hidden dimension is roughly twice that of Qwen3 and DeepSeek, its decode-stage attention is more sensitive to transfer by PCIe. \SysName's orchestration policy, which strategically offloads $OP_1$ and $OP_2$ to the CPU, effectively mitigates the transfer via PCIe, leading to the substantial improvement.

\vspace{-6pt}
\subsection{Expert Hit Ratio Comparison}
\vspace{-2pt}
Figs.~\ref{fig:eval_2}(a) and (b) show expert hit ratios of \SysName's EAS mechanism and the random selection for experts. In memory-constrained scenarios where VRAM can only hold a limited number of experts (e.g., up to 30 for DeepSeek and 50 for Qwen3), EAS achieves an expert hit ratio approximately 40\% higher than random selection. As shown in Fig.~\ref{fig:eval_2}(c), the ability to achieve a higher hit ratio directly translates to a throughput improvement of up to 1.47--1.50$\times$.

\vspace{-6pt}
\subsection{Ablation Study}
\vspace{-2pt}
Table~\ref{tab:ablation_study} shows how the components of our methods contribute to the performance. Performance was measured with Qwen3 on System (I). The most significant throughput improvement comes from coalescing expert micro-batches and performing co-execution with AMX, resulting in a 1.51$\times$ improvement. Furthermore, it is noteworthy that even method (c), which is limited to the prefill stage, provided a meaningful 1.05$\times$ improvement overall.
\section{Conclusion}
\label{section7}
\vspace{-4pt}

This paper presents \SysName, an AMX-enabled CPU–GPU collaborative system that optimizes MoE inference through a coalesced micro-batch for expert execution with an offloading strategy to enhance throughput. By incorporating coalescing-aware orchestration policy and expert-aware stratification, \SysName maximizes system utilization, achieving a 2.0$\times$ average higher throughput over SOTA methods. These results highlight \SysName's effectiveness in optimizing MoE inference for throughput-oriented workloads on resource-constrained single-GPU systems.

\section{Acknowledgement}
\label{section8}
\vspace{-4pt}

This work was supported by the Institute of Information \& Communications Technology Planning \& Evaluation (IITP) through the IITP-ITRC program (IITP-2025-RS-2020-II201847), and the IITP grant (No. RS-2025-02264029, Integration and Validation of an AI Semiconductor-Based Data Center Training and Inference System), all funded by the Korea government (MSIT).

\newpage
\bibliographystyle{ACM-Reference-Format}
\bibliography{Script/99_reference}
\footnotesize

\end{document}